# Property Classification of Vacation Rental Properties during Covid-19

Favour Yahdii Aghaebe[1,1], Dustin Foley[2,1], Prof. Eric Atwell[2,1], and Dr. Stephen Clark[2,1]

[1]Leeds Institute for Data Analytics
[2]Consumer Data Research Center
[2]University of Leeds

**Summary**

This abstract advocates for employing clustering techniques to classify vacation rental properties active during the Covid pandemic to identify inherent patterns and behaviours. The dataset, a collaboration between the ESRC funded Consumer Data Research Centre (CDRC) and AirDNA, encompasses data for over a million properties and hosts. Utilising K-means and K-medoids clustering techniques, we identify homogenous groups and their common characteristics. Our findings enhance comprehension of the intricacies of vacation rental evaluations and could potentially be utilised in the creation of targeted, cluster-specific policies.

1. **Introduction**

Travel and tourism have been embedded into our human experience for centuries. The importance of accommodation in tourism establishes relationships between the accommodation subsector and the hospitality industry. This means that changes—negative and positive—affecting the industry also affect the subsector. The Covid-19 pandemic came with a need for alterations to travel, particularly regarding public health and an increased sensitivity to security and safety (Pizam and Mansfeld, 1996).

In the United Kingdom (UK), 48% of tourist accommodation businesses cited more than 20% decline in profits following the pandemic (Office for National Statistics, 2021). As travelers and vacationers navigate these changes, concerns for safety and hygiene now take center stage (Higgins-Desbiolles, 2020). Tourists may now prioritize perceived health and safety practices. A report on hotels and restaurants (Gursoy et al., 2020) cites visible sanitizing efforts like hand sanitizers throughout the property and leaving rooms unoccupied a few days between guests among the health, safety and hygiene expectations of guests during and post pandemic.

Most vacation rental properties belonging to the peer-to-peer sharing economy rely on host-guest interactions. Hosts list and describe their properties on the website and both parties rate and review each other post stay. Hosts are also responsible for compliance with rules and regulations, although this is often low (Maria, 2015) leading to government spot checks (Barnes, 2022). Predictions show tourists favouring traditional hotels over peer-to-peer accommodation during and post pandemic because of a lack of standardization (Hossain, 2021), (Glusac, 2020).

---

[1] f.y.aghaebe@leeds.ac.uk
[2] d.j.foley2@leeds.ac.uk
[2] e.s.atwell@leeds.ac.uk
[2] s.d.clark@leeds.ac.uk

The pandemic's impact on tourism and the sharing economy has been widely researched - studies have focused on areas like: host perceptions; (Farmaki et al., 2020); impact of the pandemic on peer-to-peer accommodation pricing (Hidalgo et al., 2022); and the opinions of potential tourists on health and safety (Petruzzi and Marques, 2022). With this analysis, we aim to provide a means to augment the process of identifying suitable accommodation thereby reducing subjectivity and increasing guest confidence in the suitability of their choices.

## 2. Data and Methods

### 2.1 Data

The initial dataset consisted of over 1.3 million vacation rentals across the UK with about 526,000 corresponding host observations. The property utilisation data was presented in two levels of aggregation, daily and monthly with attributes ranging from number of bookings to a host's response rates.

To make the data suitable, data selection techniques were applied, reducing the data to about 380,000 properties and 180,000 hosts active in Great Britain during the period from the first recorded case of the pandemic on 30$^{th}$ January 2020 until the last day of restrictions – 'Freedom Day' on 19$^{th}$ July 2021.

Additionally, supplementary data derived from two indicators of population and neighbourhood characteristics —the Index of Multiple Deprivation (IMD) (Ministry of Housing, 2019), (Payne and Abel, 2012) and the Access to Health Assets and Hazards (AHAH) index (Green et al., 2018)—were incorporated into the dataset via geo-location at the Lower Layer Super Output Area (LSOA) level. Furthermore, a city-town classification at the same level was incorporated based on the House of Commons' 2018 City and Town classification (Baker, 2018).

### 2.2 Methodology

The methodology had three distinct categories: data cleaning and transformation, exploration and visualisation, and clustering. In the initial phases, the data underwent thorough cleaning to ensure accuracy and consistency. Upon inspection, the data contained several skewed variables and needed to be transformed for suitability. Data transformation involved applying a log-transformation to certain extremely skewed variables prior to the use of Principal Component Analysis (PCA) for dimensionality reduction. Exploratory analysis techniques were employed to gain insights into the dataset's characteristics and distributions using visualizations. Finally, clustering algorithms were applied to the cleaned and transformed dataset to identify groups.

#### 2.2.1 Descriptive Overview

A comparison of revenue pre and post lockdown periods is provided in **Figure 1**. When compared, the impacts the pandemic had on revenue are clearly visualized and are consistent with the results as provided by the Office for National Statistics, (Office for National Statistics, 2021).

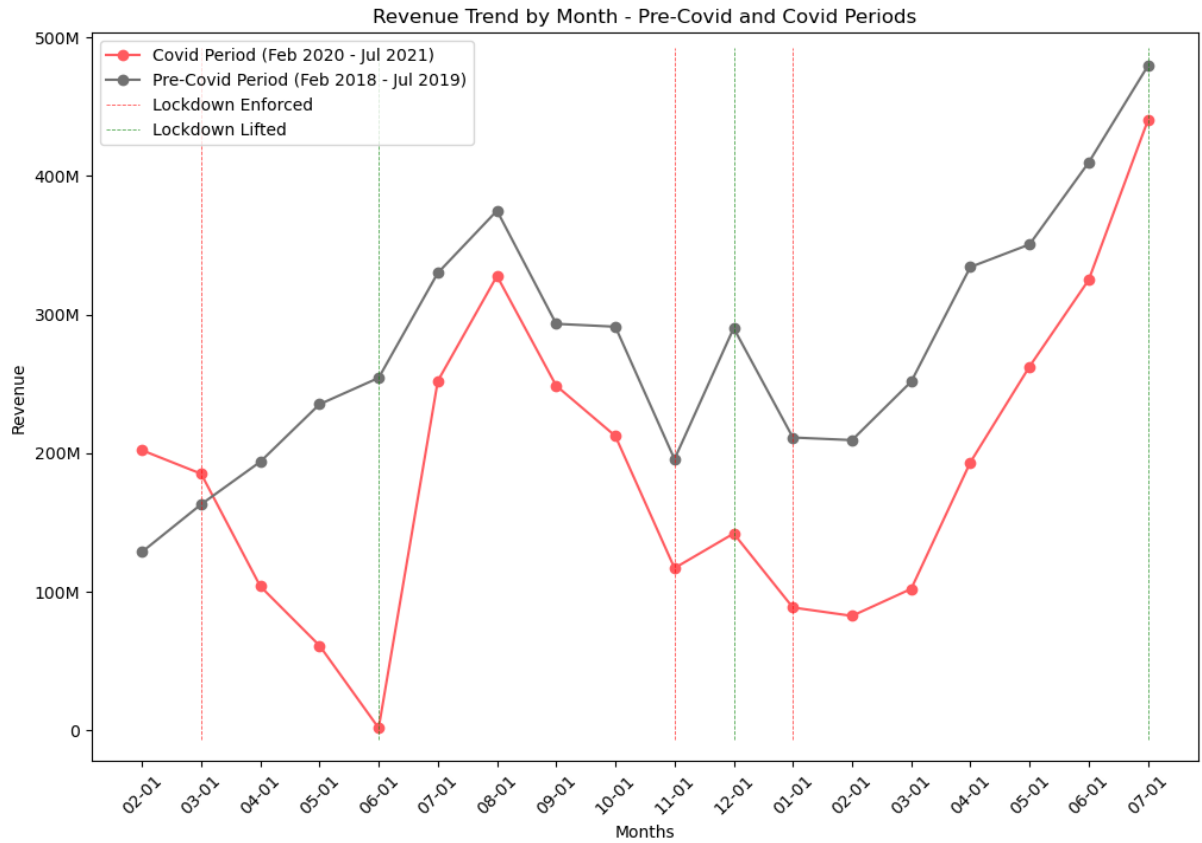

**Figure 1** Revenue Trends

**Figure 2** illustrates occupancy rates in urban and rural areas, initially appearing similar. Upon closer inspection, rural areas exhibit up to 20% higher occupancy rates, offering insights into guest behaviors during the pandemic. In 2020, especially during the first lockdown, occupancy rates were low, signaling risk aversion and stricter regulations. However, in 2021, both revenue and occupancy rates increased steadily during lockdowns, suggesting heightened guest confidence. This period may indicate individuals choosing safety-conscious vacations by booking rentals in proximity to green and blue spaces.

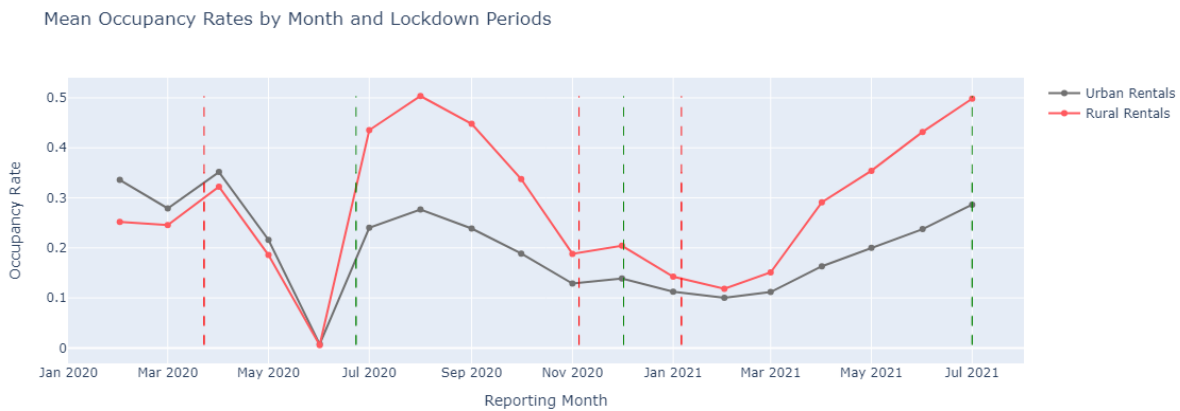

**Figure 2** Occupancy Rates (Covid Lockdown Periods)

### 2.2.2 Classification

The Scikit-learn (Pedregosa et al., 2011) implementation of the k-means and k-medoid (CLARA) algorithms were utilised for clustering. Principal Component Analysis (PCA) was used as a pre-processing step to reduce dimensionality while attempting to retain variability. The optimal number of principal components (PC) is the number for which the eigenvalue of explained variance is 1 or greater, in this case 7 as described in **Figure 3**, accounting for about 79% of data variation.

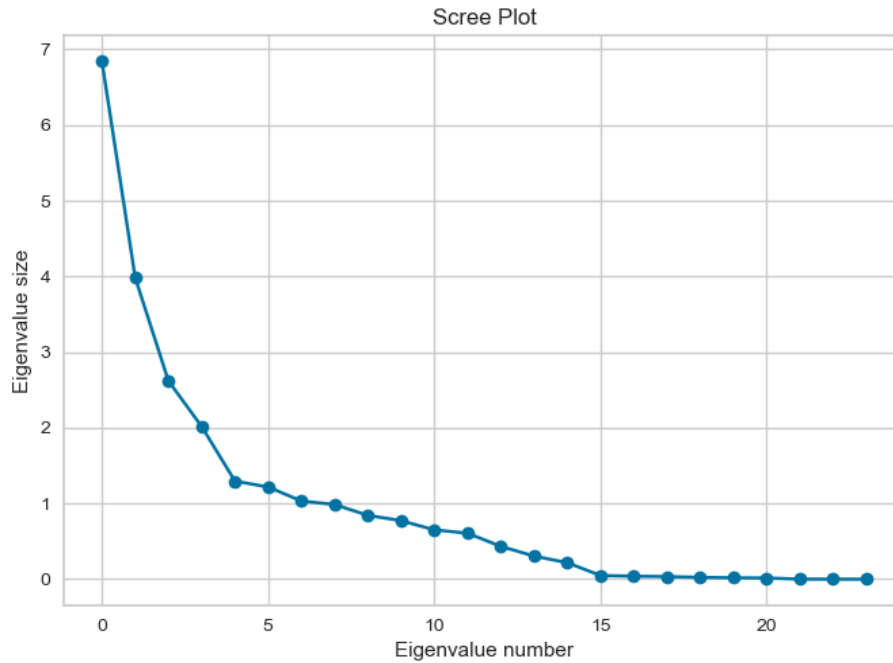

**Figure 3** PCA Explained Variance

The PC loadings for the numerical variables can be seen in **Figure 4**. We see that for example, PC 0 places a high importance on the rating variables while PC 4 loads highly on the two indexes of area based socio-economic and geographical characteristics – AHAH and IMD.

|  | PC 0 | PC 1 | PC 2 | PC 3 | PC 4 | PC 5 | PC 6 |
|---|---|---|---|---|---|---|---|
| Average Daily Rate (GBP) | -0.1098 | 0.3872 | 0.1526 | 0.0313 | -0.1910 | -0.1389 | -0.1287 |
| Annual Revenue (GBP) | -0.2148 | 0.2478 | -0.2949 | -0.0190 | -0.0807 | -0.1203 | 0.1250 |
| Occupancy Rate | -0.0519 | -0.0391 | 0.0912 | -0.0738 | -0.0001 | -0.2969 | 0.1181 |
| Number of Bookings | -0.2236 | 0.2205 | -0.3611 | -0.0482 | -0.0614 | -0.1593 | 0.1266 |
| Bedrooms | -0.0707 | 0.3071 | 0.2945 | 0.1037 | 0.1890 | 0.1678 | 0.1150 |
| Bathrooms | -0.0514 | 0.2607 | 0.2592 | 0.1309 | 0.1895 | 0.2137 | 0.0800 |
| Max Guests | -0.0710 | 0.3337 | 0.2900 | 0.1230 | 0.1084 | 0.1838 | 0.0361 |
| property_response | -0.1069 | 0.0303 | 0.0396 | -0.6397 | 0.0132 | 0.2558 | 0.0091 |
| Minimum Stay | 0.0159 | -0.0162 | 0.1244 | 0.0167 | -0.1948 | 0.0514 | 0.8113 |
| Reservation Days | -0.2060 | 0.1940 | -0.3135 | -0.0769 | -0.0891 | -0.2386 | 0.1545 |
| Available Days | -0.0257 | 0.0762 | -0.3272 | 0.2239 | 0.1625 | 0.5537 | -0.1190 |
| Blocked Days | 0.1557 | -0.1850 | 0.4494 | -0.1142 | -0.0608 | -0.2502 | -0.0147 |
| Number of Photos | -0.1115 | 0.2160 | 0.1349 | -0.0732 | -0.0685 | -0.0142 | -0.0253 |
| Overall Rating | -0.3528 | -0.1578 | 0.0862 | 0.0653 | -0.0388 | 0.0212 | -0.0564 |
| Airbnb Communication Rating | -0.3506 | -0.1634 | 0.0861 | 0.0590 | -0.0265 | 0.0259 | -0.0443 |
| Airbnb Accuracy Rating | -0.3525 | -0.1582 | 0.0839 | 0.0563 | -0.0210 | 0.0135 | -0.0480 |
| Airbnb Cleanliness Rating | -0.3505 | -0.1548 | 0.0787 | 0.0513 | -0.0176 | 0.0061 | -0.0579 |
| Airbnb Checkin Rating | -0.3514 | -0.1594 | 0.0846 | 0.0609 | -0.0254 | 0.0226 | -0.0495 |
| Airbnb Location Rating | -0.3506 | -0.1524 | 0.0855 | 0.0653 | -0.0352 | 0.0146 | -0.0639 |
| AHAH Index | 0.0785 | -0.0688 | 0.0425 | 0.1244 | -0.6224 | 0.2220 | -0.0003 |
| IMD Index | -0.0512 | 0.0383 | 0.0128 | -0.0528 | 0.5393 | -0.3237 | -0.0125 |
| Number of Listings | 0.1112 | 0.1797 | 0.0196 | -0.1172 | -0.2643 | -0.1062 | -0.4302 |
| host_response | -0.1069 | 0.0303 | 0.0396 | -0.6397 | 0.0132 | 0.2558 | 0.0091 |
| Cleaning Fee (GBP) | -0.1075 | 0.3883 | 0.1552 | 0.0335 | -0.1893 | -0.1339 | -0.1308 |

**Figure 4** PC Loadings

The next step after dimensionality reduction involved encoding categorical variables using ordinal encoding for ordinal variables and one hot encoding for nominal variables and then utilising the 'elbow' method to identify an optimal K value. The elbow method is based on the intuition in (Satopaa et al., 2011) which provides some scientific objectivity and balance to the art of hyperparameter tuning. The resulting elbow plots are presented in **Figure 5** below.

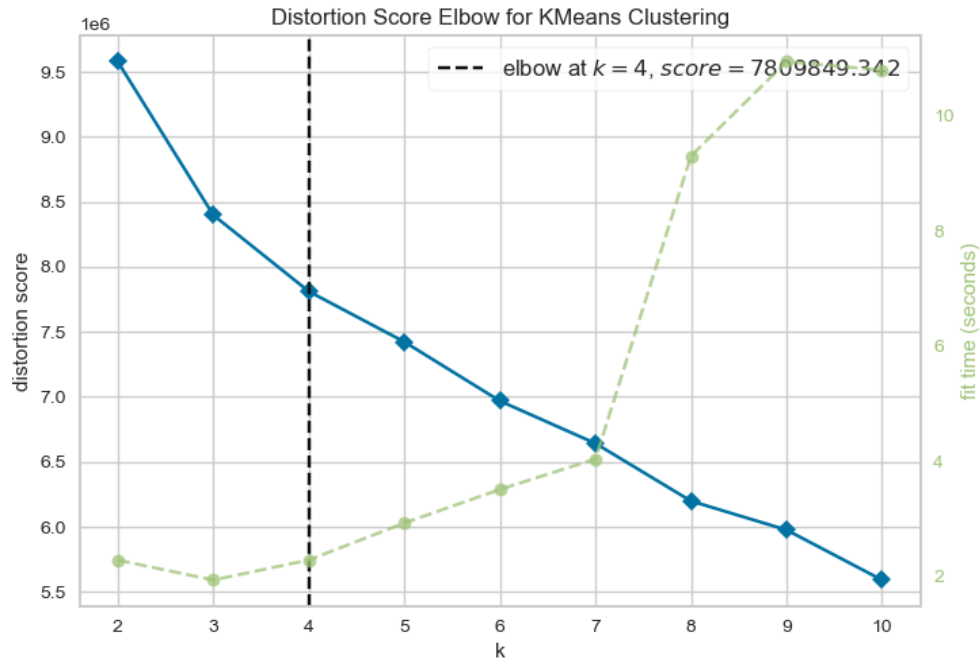

**Figure 5 (a)** Elbow Plot: K-means Algorithm

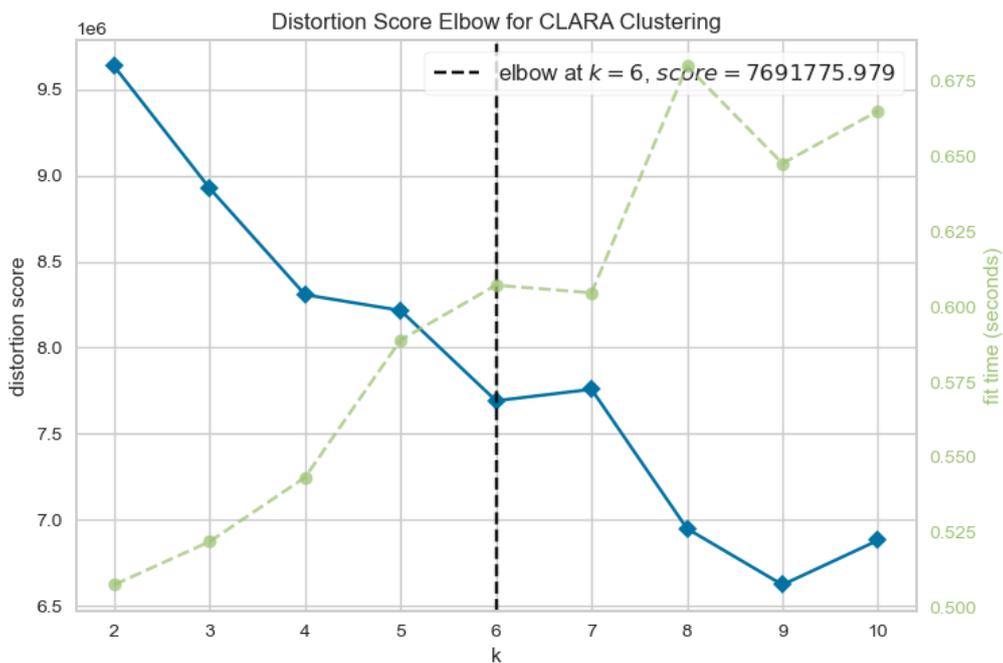

**Figure 5 (b)** Elbow Plot: K-medoids Algorithm

## 3. Results

Notable differences can be seen in the group and urban-rural distribution of both algorithms.

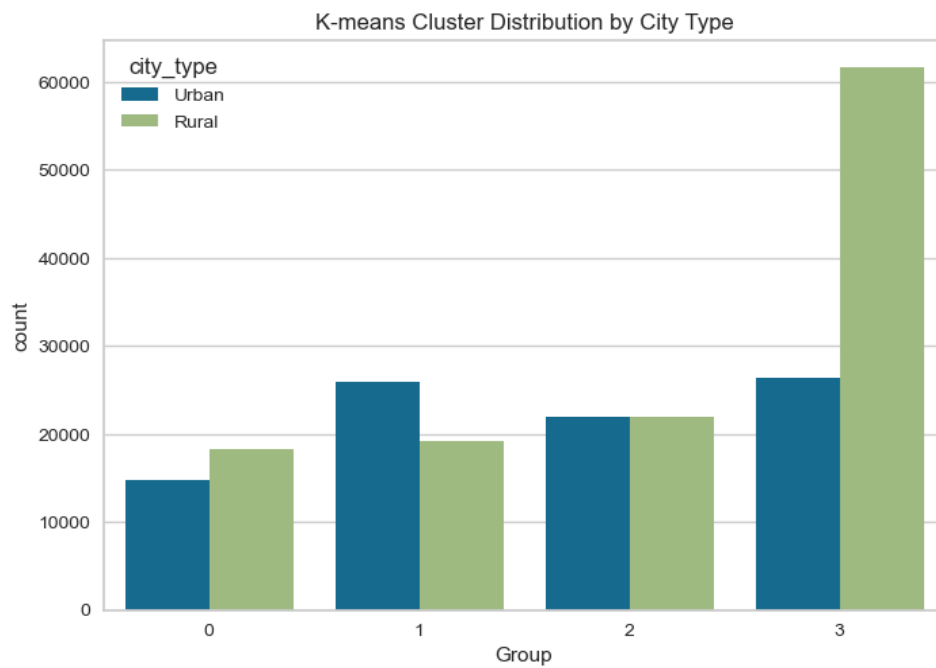

**Figure 6(a)** Property Distribution within Groups (Urban and Rural Areas): K-means

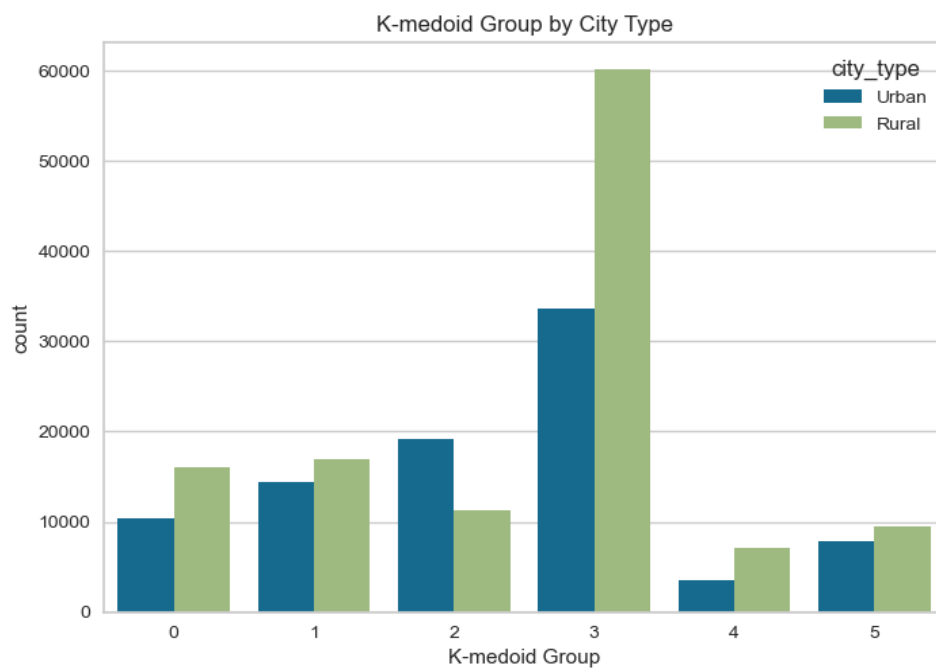

**Figure 6(b)** Property Distribution within Groups (Urban and Rural Areas): K-medoids

In **Table 1** below, we provide a cross tabulation of the results of cluster membership from both algorithms. The question of evaluation and deciding which algorithm has performed 'best' then comes into play. This is addressed in section 3.1 below.

| K-medoids/ K-means | 0 | 1 | 2 | 3 | 4 | 5 | Total K-means |
|---|---|---|---|---|---|---|---|
| 0 | 1752 | 67499 | 25 | 386 | 11 | 39 | **69712** |
| 1 | 216 | 488 | 67776 | 16917 | 8354 | 737 | **94488** |
| 2 | 32782 | 179 | 1517 | 31458 | 2514 | 37945 | **106395** |
| 3 | 12556 | 3 | 91 | 98861 | 2907 | 1318 | **115736** |
| **Total K-medoids** | 47306 | 68169 | 69409 | 147622 | 13786 | 40039 | 386331 |

**Table 1** Property Distribution within Clusters: K-medoids Vs. K-means

### 3.1 Cluster Evaluation and Comparison

Having utilised more than one algorithm for classification in this project both resulting in different cluster distributions, as seen in **Table 1** above, the next natural thought that comes to mind is that of identifying which clustering technique has performed 'best'. This task of identifying which algorithm has provided the best classifications in the absence of labelled ground truth data is one that is heavily reliant on domain knowledge and subjectivity. As highlighted by (Ackerman et al., 2010), there is no consensus on what good clustering looks like. However, there exist some internal methods of validation that use intrinsic characteristics of the dataset to identify optimal classification. Ideally, these should be used in tandem with external methods of validation – involving true and predicted classes. For this project, we have utilised two of such internal methods.

Method 1: Davies-Bouldin Index (DBI)

The DBI (Davies and Bouldin, 1979) is an internal metric for cluster evaluation, it works by calculating the average similarity of each cluster with a cluster close to it. Similarity between clusters is computed as a function of inter and intra cluster dispersion. The index is calculated in a three-part process: first the intra cluster dispersion is calculated then the inter cluster dispersion/separation measure is calculated and finally the similarity between both dispersions is calculated. "The aim of this metric is to minimize the similarity calculated between all pairs of clusters. The Scikit-learn implementation of this index was used, and the following scores in **Table 2** below were realised:

| Model | DBI Score |
|---|---|
| **K-Means** | 1.98 |
| **K-Medoids** | 2.06 |

**Table 2:** DBI Score for K-Means and K-medoids.

Method 2: Calinski-Harabasz Index (CHI)

The CHI (Caliński and Ja, 1974), also referred to as the variance ratio criterion is calculated as the sum of intra and inter cluster dispersion for all clusters where dispersion is the sum of squared distances. The aim is to maximise the CHI score as this is a good indicator of denseness within a cluster and separation between clusters. First, the inter-cluster dispersion is calculated, then the intra-cluster dispersion is calculated, and finally CHI is computed as the sum of these dispersions. The Scikit-learn implementation of this index was used, and the following scores in **Table 3** below were realised:

| Model | CHI Score |
|---|---|
| **K-Means** | 68703.70 |
| **K-Medoids** | 43040.85 |

**Table 2:** CHI Score for K-Means and K-medoids.

The outcomes of the aforementioned metrics guided the adoption of the results from the K-Means algorithm, and subsequently, the creation of cluster profiles.

**3.2 Profiling**
As with any clustering analysis, the goal is classification to potentially profile items for some purpose or another, commercial or otherwise. The results of the host and property classification in this study could be utilised to create targeted commercial campaigns towards specific groups of hosts as well as contribute towards the creation of targeted government policies with the aim of reducing host non-compliance and the effective implementation of the UK governments hospitality strategy (Gov.UK, 2023).

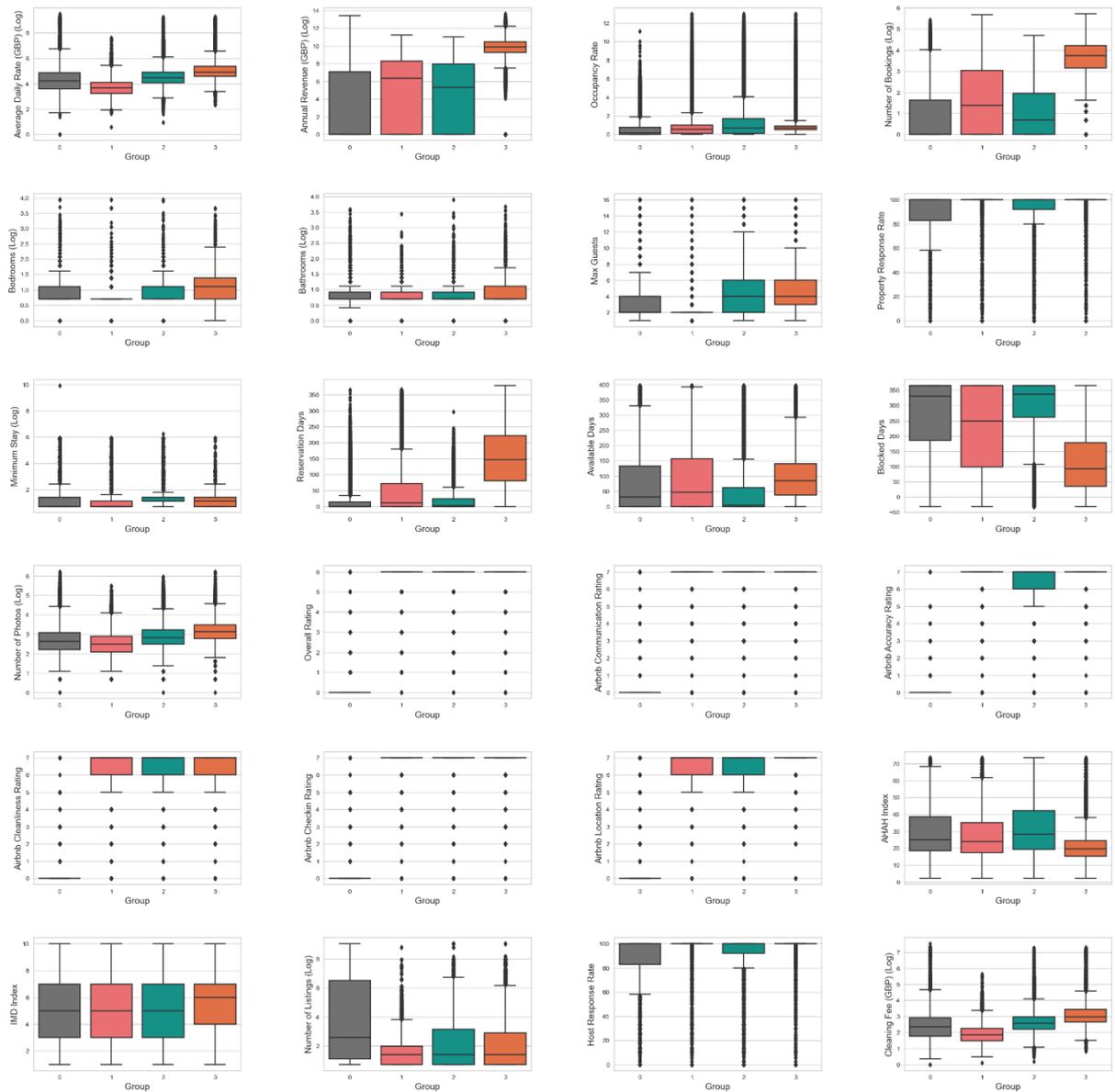

**Figure 7** Inter-Cluster Distinctions (a)

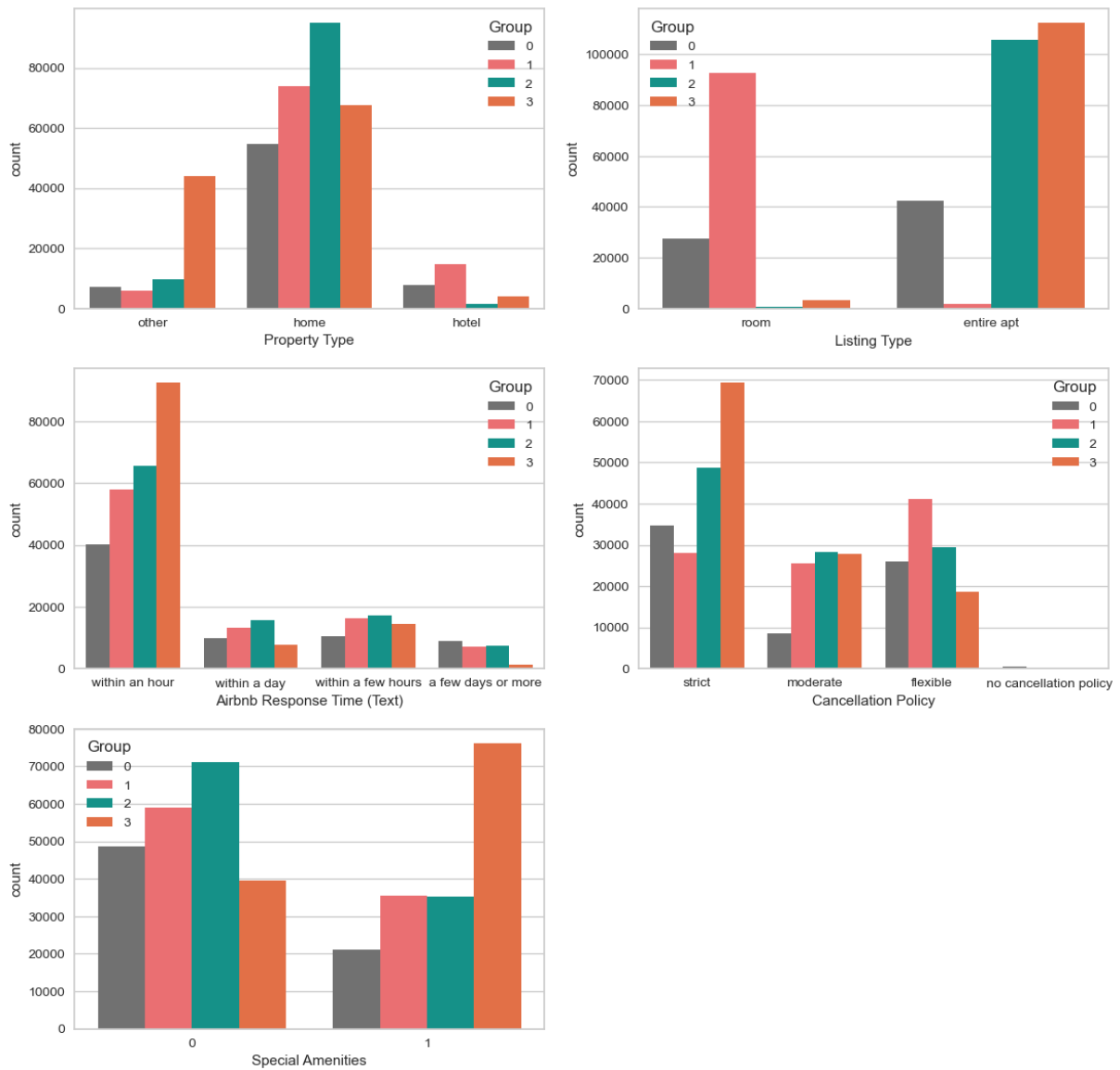

**Figure 7** Inter-Cluster distinctions (b)

Based on the characteristics of the clusters presented in **Figure** 6 and **7** above, the following profiles have been created for these groups of properties and hosts:
- Group 0: These properties feature hosts with poor ratings.
- Group 1: These properties are hotels and single rooms available within homes or flats.
- Group 2: These properties comprise majorly of properties that were immensely popular during the period in review.
- Group 3: These were predominantly commercial rural properties.

A spatial representation of cluster distribution in Leeds City Centre is presented in **Figure 8** below. From this, we can see that properties in group 0 cluster around the City Centre more than the properties from other groups.

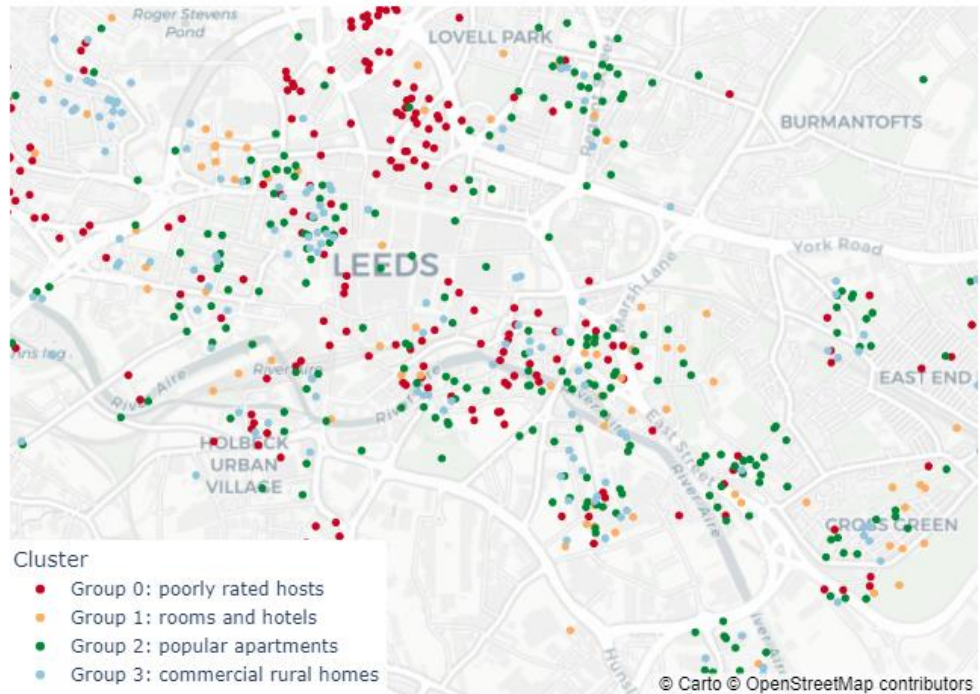

**Figure 8:** Spatial representation of Property Clusters in Leeds

4. **Conclusion**

In conclusion, this study illuminates the vacation rental subsector's dynamics during the COVID-19 pandemic, employing clustering algorithms for insightful property classification and providing actionable insights for stakeholders in strategic decision-making. The results offer objectivity for potential guests in selecting vacation rentals and furnish policymakers with cluster-specific information to formulate effective regulations. While limited to Great Britain, this study suggests opportunities for broader research, including extending the study to different locations and encompassing properties beyond the sharing economy.

5. **Acknowledgements**

The data was provided by the Consumer Data Research Centre, an ESRC data investment, under project ID ESRC grant ES/L011891/1 in conjunction with AirDNA.

**Biographies**

**Favour Aghaebe** is a Data Scientist and Early Career Researcher at the Leeds Institute for Data Analytics, University of Leeds. Her research interests include consumer behaviour and health care.

**Dustin Foley** is a Research Software Engineer for the Consumer Data Research Centre (CDRC) at the University of Leeds. He often works with datasets involving urban mobility or natural language processing, and previously conducted research for the University's School of Law and School of Food Science & Nutrition.

**Stephen Clark** is a Research Fellow in the Consumer Data Research Centre, University of Leeds. His research interests include the areas of politics, health, retail, and housing.